# Fractional Multiscale Fusion-based De-hazing

Uche A. Nnolim[1]

*Department of Electronic Engineering, University of Nigeria, Nsukka, Enugu, Nigeria*

**Abstract**

*This report presents the results of a proposed multi-scale fusion-based single image de-hazing algorithm, which can also be used for underwater image enhancement. Furthermore, the algorithm was designed for very fast operation and minimal run-time. The proposed scheme is the faster than existing algorithms for both de-hazing and underwater image enhancement and amenable to digital hardware implementation. Results indicate mostly consistent and good results for both categories of images when compared with other algorithms from the literature.*

*Keywords:* Fractional order calculus-based multi-scale contrast operator; hybrid local-global contrast enhancement; underwater image enhancement processing; hazy image contrast enhancement; entropy guided fusion

## 1. Introduction

There are numerous algorithms for hazy and underwater images since both image groups share similar degradations such as low contrast, poor visibility in effect of suspended solids in fluids (air or water). These are primarily classified as restoration and enhancement-based approaches. Furthermore, a third category is the fusion-based methods, which is similar to a hybrid approach. Recently, there have been proposed algorithms involving deep-learning and convolutional neural networks for both underwater and hazy image processing. However, these are all usually highly complex approaches and are not normally suited to real-time operation or hardware implementation. Furthermore, such algorithms increasingly require a high degree of computational resources for operation. Additionally, results are not always commensurate to the level of complexity involved in the approaches. Thus, the proposed scheme aims to mitigate the issue of complexity, while improving or maintaining effectiveness.

## 2. Background

For the hazy images the most popular restoration-based approach is the dark channel prior (DCP) de-hazing method by He et al [1], which has been modified in numerous ways and variants. A review of the DCP-based approaches can be found in work by Lee et al [2]. Others include [3] [4] [5] [6] [7] [8] [9] [10] [11] [7] [12] [4] [13] [14] [15] [16]. The enhancement-based methods include [17] [18] [19] [20] [3] [4] [5] [21] [22] [23]. More recent works include Artificial Multiple-Exposure Image Fusion (AMEF) de-hazing algorithm by Galdran [24] and deep learning and convolutional neural network architectures, which include works by [25] [26] [27] [28] [29] [30] [31] [32] [33] [34] [35] [36] [37] [38] [39] [40] [41] [42] [43].

Additionally, for underwater image enhancement popular  methods include those by Galdran et al [44], Li et al [45], Li and Guo [46], Zhao et al [47], Chiang and Chen [48], Wen et al [49], Serikawa and Lu [50], Carlevaris-Bianco et al [51], Chiang et al [52], etc. Also the enhancement-based approaches include those by Iqbal et al [53], Ghani and Isa [54], Fu et al [55], Gouinaud et al [56], Bazeille et al [57], Chambah et al [58], Torres-Mendez and Dudek [59], Ahlen et al [60] [61], Petit et al [62], Bianco et al [63], Prabhakar et al [64], Lu et al [65], Li et al [45] and Nnolim [66] [67]. Other algorithms include those by Prabhakar et al [64], Garcia et al [68], Rzhanov et al [69], Singh et al [70] and Fu et al [55].

## 3. Proposed algorithm

The proposed scheme combines a multi-scale fractional filtering operation with a decision-based scheme using entropy and standard deviation to maximize contrast enhancement. The multiscale structure decomposes the detail and approximation coefficients, enhancing the former while reducing the latter. This is performed in fractional calculus domain, which has advantages over conventional integer-order calculus. The details of the proposed algorithm [71] is shown in Fig.1.

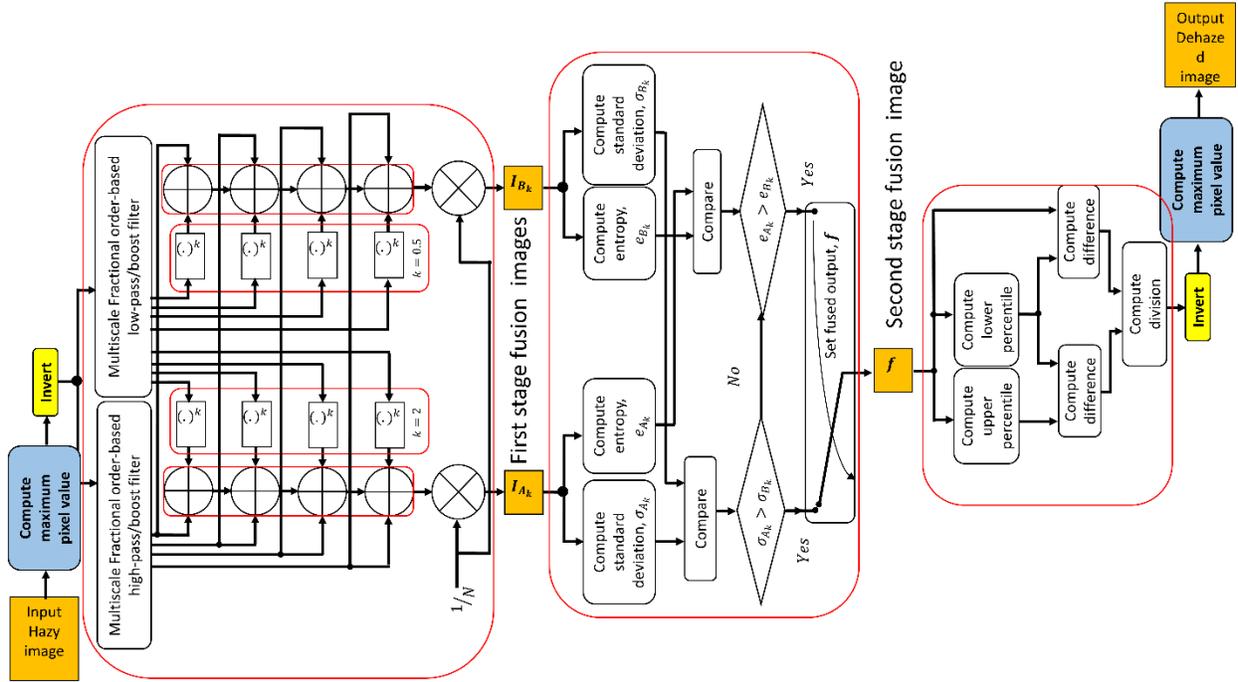

**Fig. 1.** Proposed algorithm (PA) for enhancing hazy and underwater images

### 3.1 Preliminary results

The proposed scheme can be utilized in fractional high-pass filter (HPFC) or high-boost filter (HBFC) configurations. The former configuration leads to sharper edges but darker images, while the latter results in brighter images with moderately sharpened edges. In Fig. 2, a sample result of the algorithm is shown for high-pass and high-boost configurations.

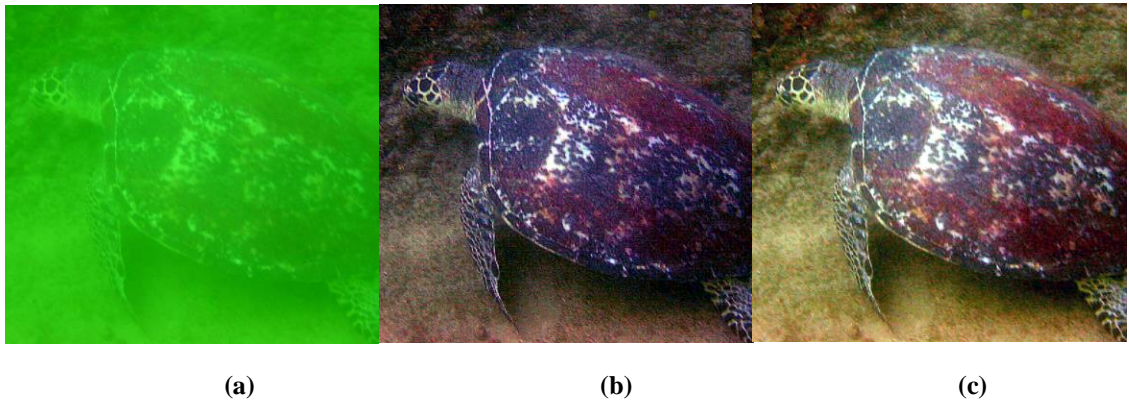

**(a)**        **(b)**        **(c)**

**Fig. 2.** (a) Underwater image enhanced with (b) PA using high-pass and (c) high-boost fractional filter settings

## 4.  Experiments

Results are presented in Fig. 3, which contains results from [45], amended with results from [67] and PA and show that there is a considerable contrast and edge enhancement as details are seen much more clearly with minimal haze. For the *fish* image in Fig. 3, only results by Ancuti et al [72], Fu et al [55], Galdran et al [44], PDE-based PWL-CLAHE (forward and reverse configuration) [67] and PA yield good results. The rest of the image results depict hazy, faded images with large degree of green colouration, while Li et al's method [45] yields an image with reddish colouration, implying over-compensation of red channel in the processed image.

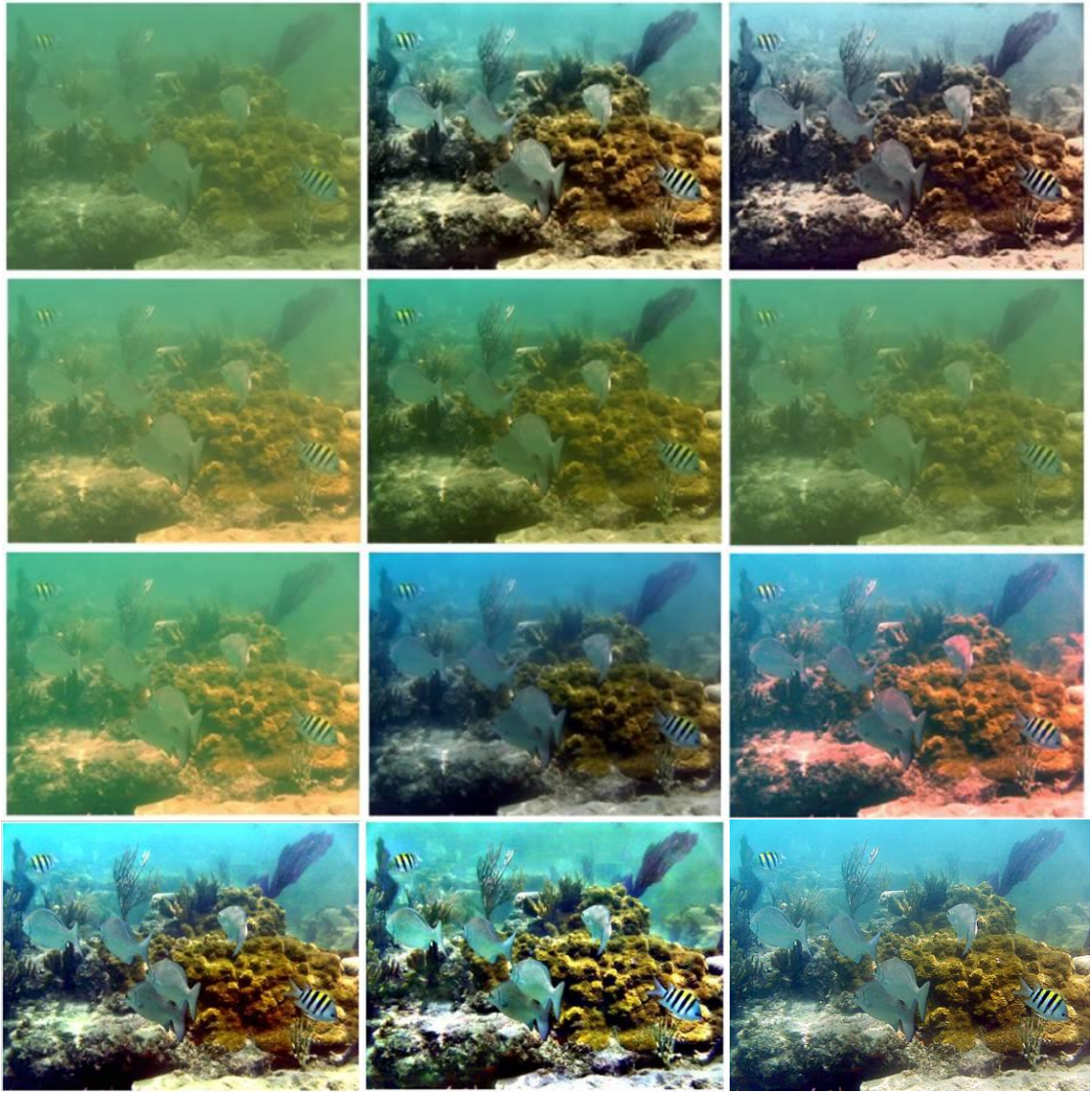

**KEY**

| (a) | (b) | (c) |
|-----|-----|-----|
| (d) | (e) | (f) |
| (g) | (h) | (i) |
| (j) | (k) | (l) |

**Fig. 3.** Figure from [45], amended with results from [67] and PA (a) Original *Fish2* image processed with algorithms proposed by (b) Ancuti *et al* [72] (c) Fu *et al* [55] (d) Chiang and Chen [48] (e) He *et al* [1] (f) Carlevaris-Bianco *et al* [51] (g) Serikawa and Lu [50] (h) Galdran *et al* [44] and (i) Li *et al* [45] (j) PWL-CLAHE and (k) CLAHE-PWL-AD configurations (l) PA

Based on the results, the proposed algorithm yields finer, sharper edges and details with minimal intrinsic noise due to the fractional derivative ability. The visual results are mostly reflected in the quantitative metrics shown in Table 1, with PA showing the highest AG values, indicating more visible edges and details especially on the rock face of the bottom left corner of the image (image(l)). However, PDE-GOC2-CLAHE yields the highest colourfulness (C) [73] and entropy while the method by Fu, et al gives best GCF value [74] (though there is over-exposure in the bright regions of the rock faces in image (c)).

**Table 1.** Comparison of PA with various algorithms for *Fish2* image

| Measures \Algos | Ancuti [72] | Bianco [51] | Chiang [48] | Fu [55] | Galdran [44] | He [1] | Li [45] | Serikawa [50] | (PDE-PWL-CLAHE) [67] | PA |
|---|---|---|---|---|---|---|---|---|---|---|
| **Entropy** | 7.8438 | 7.1251 | 7.2986 | 7.8628 | 7.6376 | 7.4587 | 7.7168 | 7.4531 | **7.8945** | 7.2558 |
| **GCF** | 9.5759 | 4.6944 | 3.9611 | **9.6404** | 8.7299 | 6.372 | 7.0632 | 4.9016 | 8.6257 | 6.9014 |
| **C** | 54.5704 | 42.3128 | 54.8975 | 36.337 | 64.0309 | 57.0533 | 63.8498 | 63.2207 | **77.5420** | 61.1957 |
| **AG** | 9.1638 | 4.1501 | 4.1285 | 9.4732 | 5.6937 | 5.174 | 7.6573 | 5.2034 | 10.4343 | **13.9107** |

In Fig. 4(1), PA yields similar results (though with much sharper edges and details) to methods by Iqbal and Ancuti et al [72], which yield best results both subjectively and quantitatively in the work by Yang and Sowmya [75].

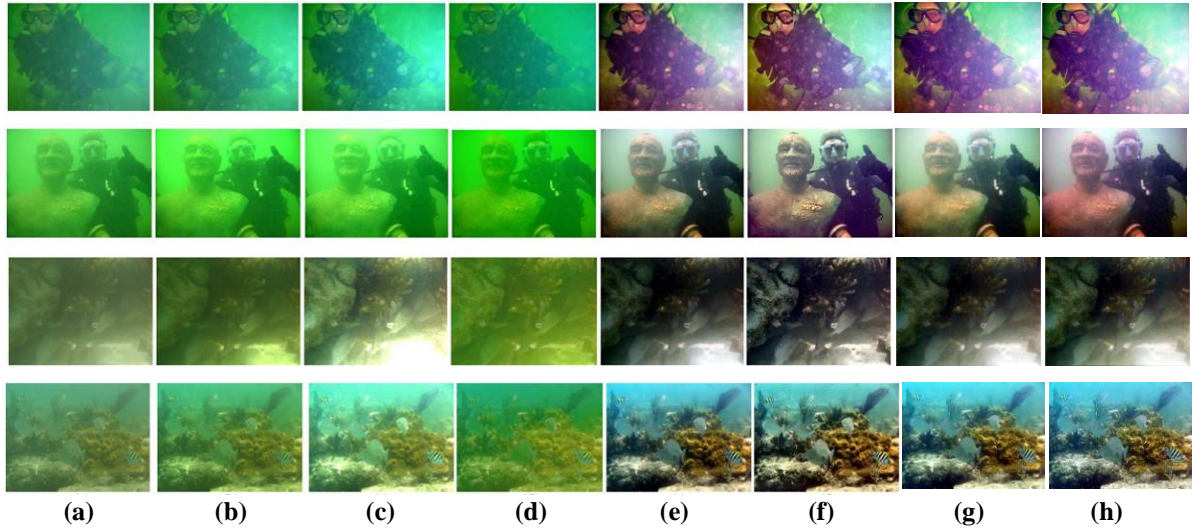

**Fig. 4.** Figure from Yang and Sowmya [75] amended with PA (a) Original images (b) He et al (c) Fattal et al (d) Tarel et al (e) Iqbal et al (f) Ancuti et al (g) PA (HBFC) (h) PA (HPFC)

**Table 2.** Comparison of PA with various algorithms from (a) Yang and Sowmya [75] (b) & (c) Emberton et al [76]

| Algos [75] / Measures | He et al | Fattal | Tarel et al | Iqbal et al | Ancuti et al | PA (HBFC) | PA (HPFC) |
|---|---|---|---|---|---|---|---|
| **CEF** [73] | 1.2461/ 1.3997/ 1.7631/ 1.4651 | **1.4884**/ 1.2338/ 1.6396/ 1.8201 | 1.4204/ **1.5814**/ **2.2850**/ 1.8838 | 0.8304/ 0.5785/ 0.6372/ 1.8447 | 1.0733/ 0.6542/ 0.7347/ 1.5900 | 1.0729/ 0.7474/ 0.8519/ **2.0880** | 0.8791 0.6261/ 0.5079 1.6936 |
| **UIQM** [77] | 2.9156/ 2.9680/ 3.0227/ 2.3403 | 3.0770/ 2.9539/ 2.7092/ 2.9490 | **4.5862**/ 3.5435/ 3.1818/ 3.2217 | 3.0701/ 3.0145/ 3.6372/ 3.4328 | 2.7267/ 3.0207/ 3.6981/ 3.6369 | 2.6546/ 3.5242/ 3.1641/ 4.0026 | 3.7747 **4.4788**/ **4.0996**/ **4.4907** |
| **UCIQE** [75] | 0.5874/ 0.5228/ 0.4818/ 0.5631 | 0.6448/ 0.5439/ 0.6158/ 0.6503 | 0.5821/ 0.5046/ 0.5308/ 0.5828 | 0.7684/ 0.6797/ 0.5919/ 0.7507 | **0.8937**/ **0.8551**/ 0.7441/ **0.8814** | 0.7511/ 0.7961/ 0.8768/ 0.8717 | 0.8014 0.6829/ **0.9034** 0.6878 |

However, the CEF gives counter-intuitive results where Iqbal, Ancuti and PA yield lower colourful values compared to methods by He, Fattal and Tarel. Nonetheless, we infer that the PA performs better in terms in subjective experiments since the method by Ancuti et al and Iqbal et al yield best results in terms of UCIQE.

We also compare PA with the algorithms and figure from [43] in Fig. 5 and though PA does not yield the best visual result for all images, it performs comparatively well.

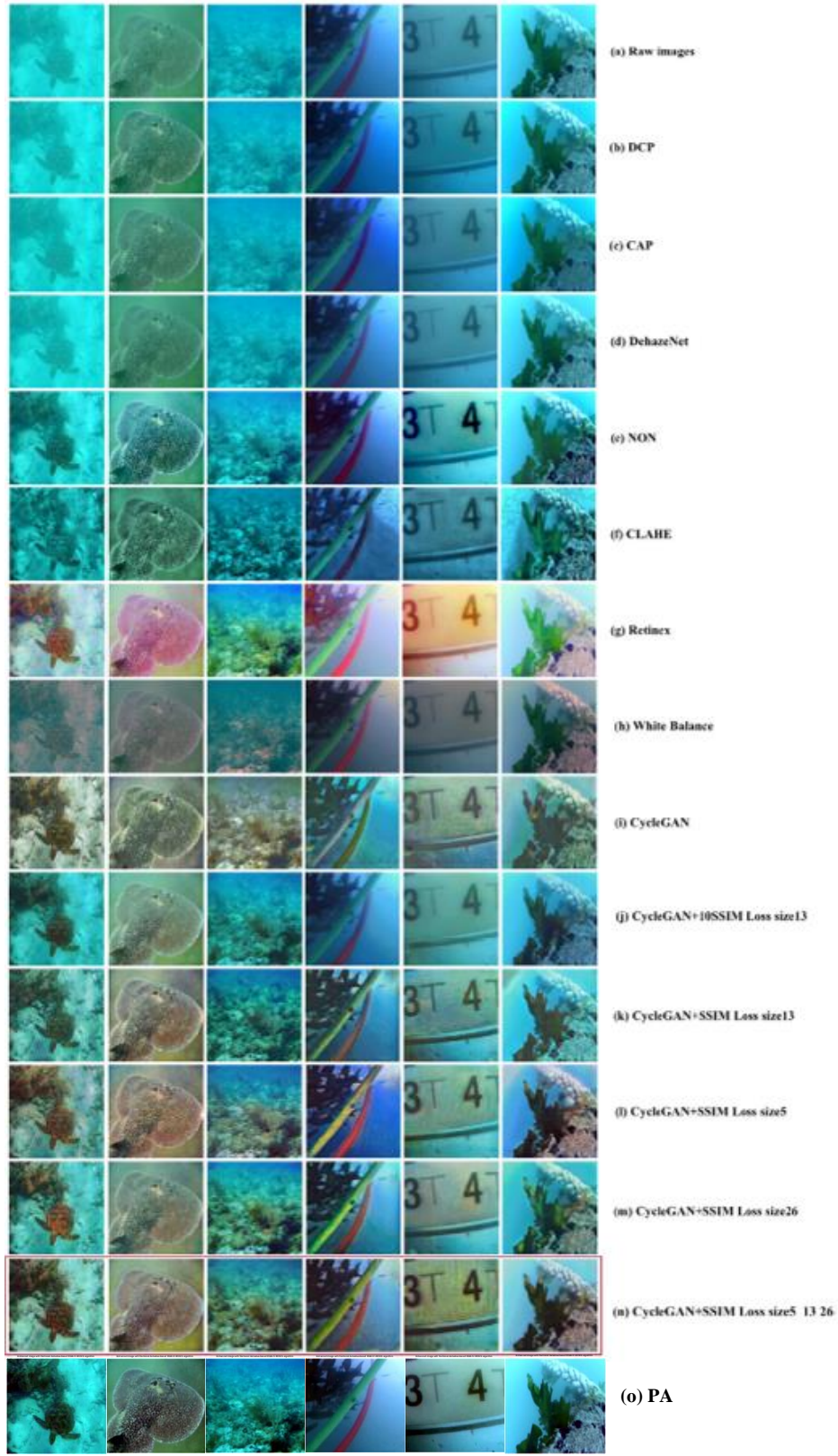

**Figure 5.** Figure from [43] amended with visual result of PA for comparison

The algorithms listed in [43] include Contrast Limited Adaptive Histogram Equalization (CLAHE), Retinex, White balance, combined Methods of He et al, Zhu et al and Non-local de-hazing, deep neural network-based schemes such as DehazeNet, cycle-consistent adversarial networks (CycleGAN), Li's method and the adjusted CycleGAN to compute structural similarity index metric (SSIM) loss (CycleGAN+SSIM Loss size). It should be noted that several of these algorithms possess high computational and structural complexity. Furthermore, since no run-time is provided for several of these algorithms, it is difficult to compare their execution time.

The proposed approach (PA) is much more vivid as it enhances edges and avoids discolouration of sky regions as seen in the *Tiananmen* image in Fig. 6. Best results are observed for PA, Ren, et al [16], Zhu, et al [78] and He et al [1] (has halos) followed by PDE-GOC-SSR-CLAHE [22] (has some halos) and PDE-IRCES [23] (no halos but under-enhanced in some regions). The method by Tarel and Hautiere [79] shows over-enhancement of edges and discolouration of sky region similarly to PDE-IRCES. The method by Ren, et al shows sharpened features without sky discolouration or over-enhancement similar to Zhu, et al (which is darker). The PDE-GOC-SSR-CLAHE yields considerable detail in non-homogeneous regions, while PA yields the highest detail and edge enhancement without sky discolouration or halo effects. Ju et al, yields excessive enhancement of contrast and colour distortion. AMEF yields a dull image and increasing the clip limit will yield similar results to Ju et al.

The same is observed for the *toys* image in Fig. 7 as the image obtained from PA has the most enhanced edges and details compared to the other results. The PDE-GOC-SSR-CLAHE gives best local contrast enhancement, followed by the DCP method by He et al, and the methods by Wang, et al [80], Ren et al, Dai, et al [81] and Zhu, et al. The rest of the other image results are faded and still contain a reasonable amount of haze or have colour distortion or saturation with minimal edge enhancement. AMEF yields a cloudy image with halos clearly visible, while PA enhances edges with adequate de-hazing and no halos.

For the *mountain* image in Fig. 8, the method by Tarel and Hautiere yields over-enhanced image, while Wang et al (more halos) and PDE-Retinex (less halos) yield bright images. The same with PDE-IRCES (no halos) and Galdran et al (more retained fog), while PA subtly de-hazes the image with some brightness to avoid darkening as this is one of the more difficult images.

For the *canyon* image in Fig. 9, Ren et al, Zhu e tal, He et al, PDE-Retinex, PA and PDE-IRCES yield best contrast. Method by Nishino et al yields over-enhanced image with artificial look. The method by Fattal and Tan indicate moderate de-hazing.

For the *pumpkins* image in Fig. 10, there are generally good results in most cases in terms of visibility improvement. However, Ren et al and Dong et al, Fattal, He et al, and PDE-Retinex give best results. This is followed by Zhu et al, Yeh et al, Ancuti et al, Kratz and Nishino, PDE-IRCES and PA. However, Ren, Zhu, He, PDE-Retinex, PA and PDE-IRCES yield best contrast (in terms of objective metrics in Table 3).

For the *city1* image in Fig. 11, Tarel et al and Nishino et al yield over-enhanced images. The method by Galdran et al and Dai et al yield images with faded colours and near-white background in the upper, sky region. He et al (visible halos and darkened regions) and PDE-Retinex yield images with enhanced sky regions. PDE-IRCES and PA yield adequately sharpened images with minimally enhanced sky regions with no halos. Ren et al and Wang and He yield images with burnt appearance, while Zhu et al gives a dark image.

For the *train* image in Fig. 12, Most results are generally good. However, Ren et al and Zhu et al yield images with poor visibility restoration. The PDE-Retinex, PDE-IRCES and PA give the highest contrast enhancement and RAG value in Table 3.

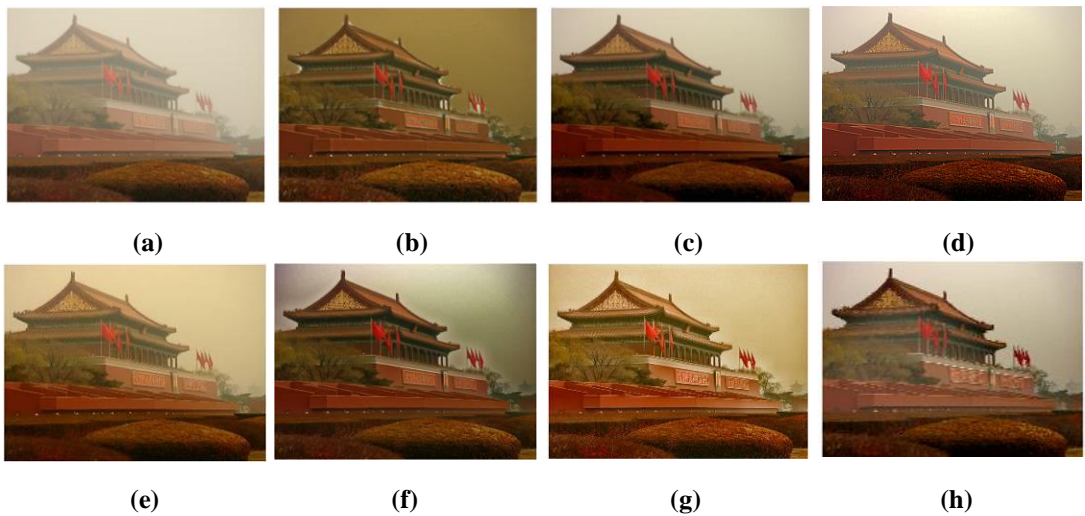

(a)　　　　　　(b)　　　　　　(c)　　　　　　(d)

(e)　　　　　　(f)　　　　　　(g)　　　　　　(h)

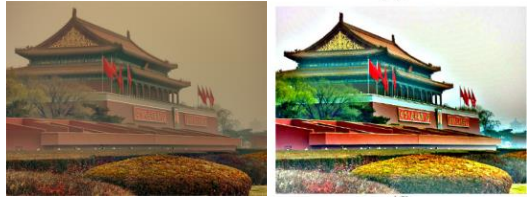

**(i)**              **(j)**

**Figure 6.** (a) Original hazy image (b) Tarel, et al (c) Zhu, et al (d) PA (e) PDE-IRCES (f) He, et al (g) PDE-GOC-SSR-CLAHE (h) Ren et al (i) AMEF (j) Ju et al

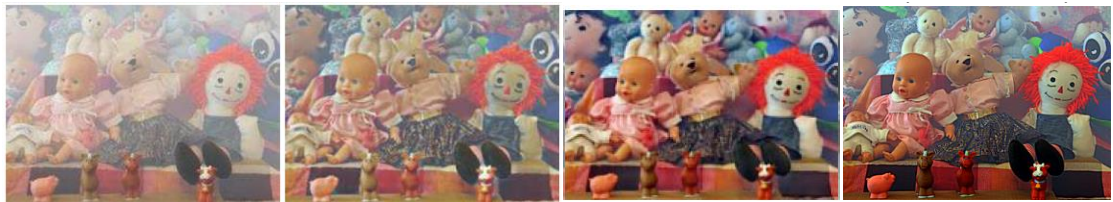

**(a)**         **(b)**         **(c)**         **(d)**

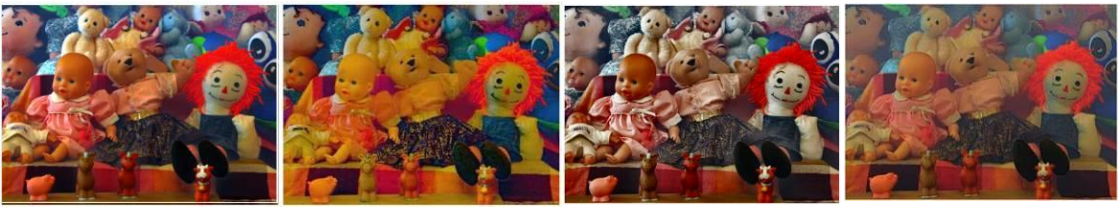

**(e)**         **(f)**         **(g)**         **(h)**

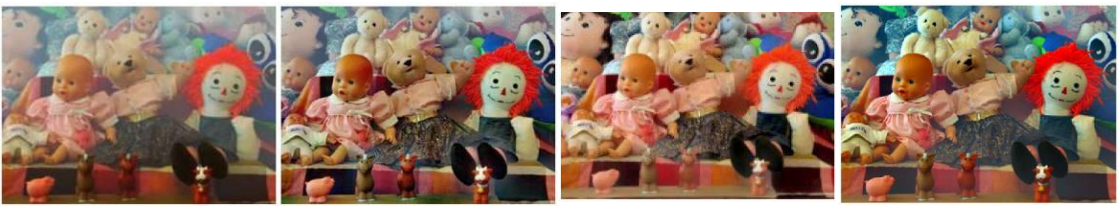

**(i)**         **(j)**         **(k)**         **(l)**

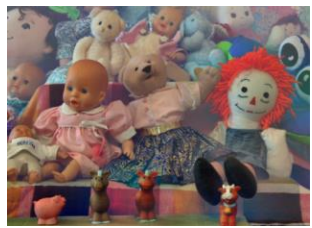

**(m)**

**Figure 7.** (a) Original hazy image (b) Tarel, et al [79] (c) Dai et al [81] (d) PA (e) He et al [1] (f) Nishino, et al [82] (g) PDE-GOC-SSR-CLAHE [22] (h) PDE-IRCES [23] (i) Galdran, et al (EVID) [11] (j) Wang & He [80]  (k) Zhu, et al [78] (l) Ren, et al [16] (m) AMEF

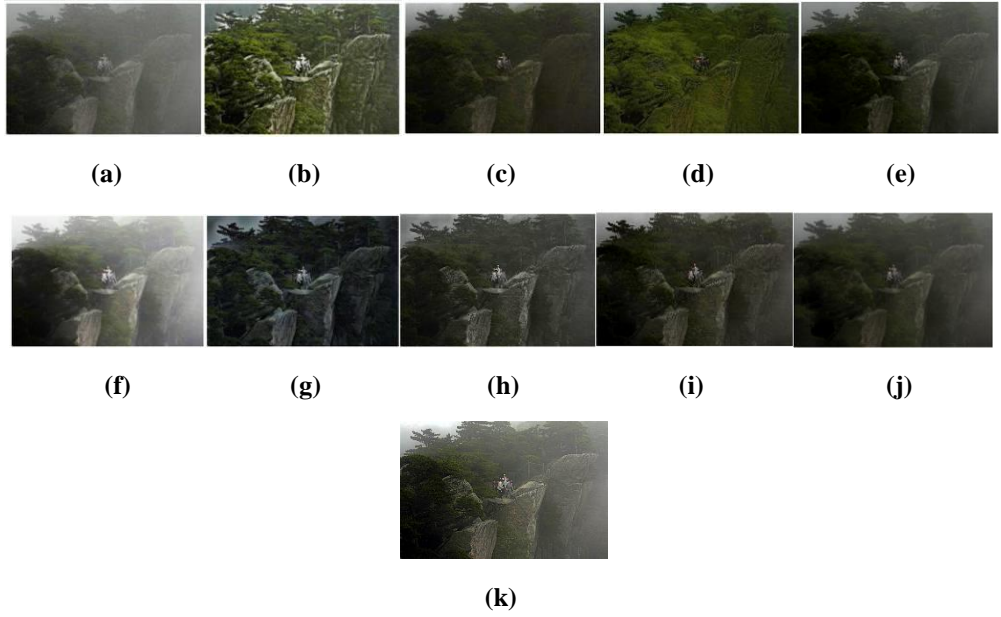

**Figure. 8** (a) Original hazy image (b) Tarel et al (c) He et al (d) Nishino et al (e) PDE-IRCES (f) Galdran et al (g) Wang & He (h) PDE-Retinex (i) Ren et al (j) Zhu et al (k) PA

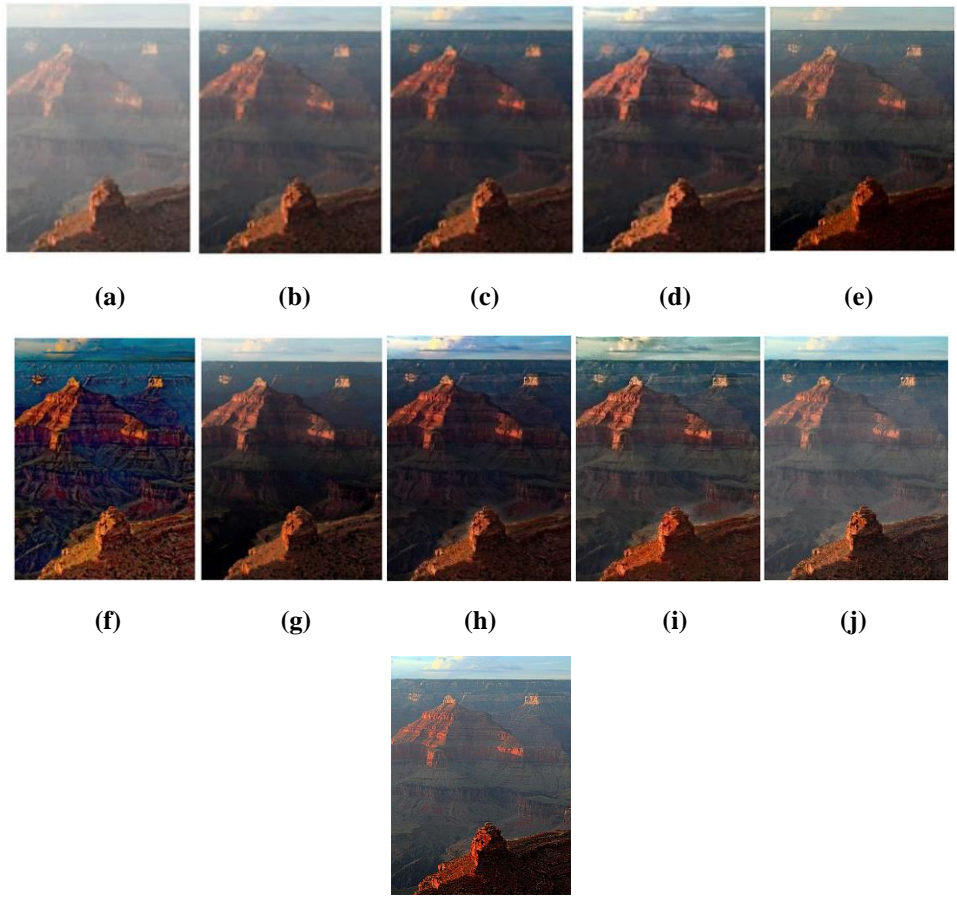

**Figure. 9** (a) Original hazy image (b) Fattal (c) Tan (d) Yang et al (e) PDE-IRCES (f) Nishino et al (g) Zhu et al (h) He et al (i) PDE-Retinex (j) Ren et al (k) PA

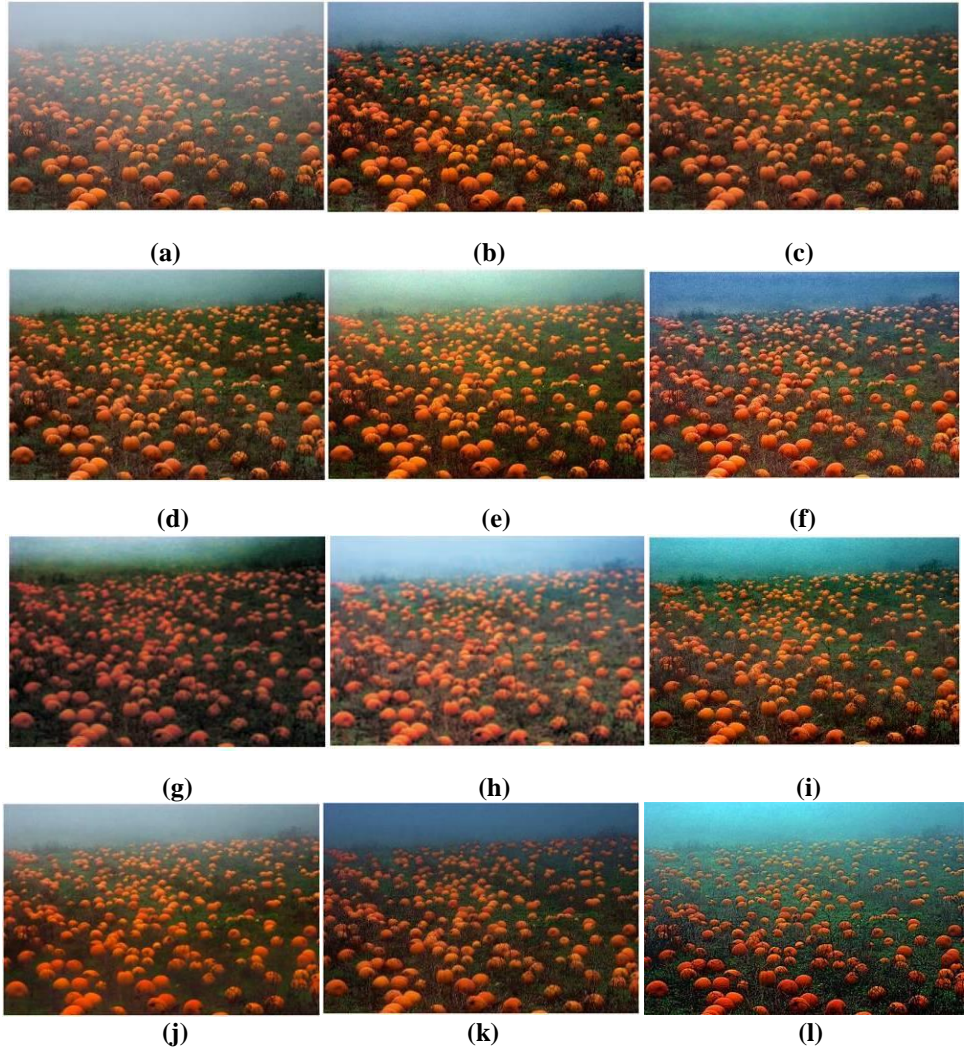

**Figure. 10** (a) Original hazy image (b) Fattal (c) Dong et al (d) He et al (e) Yeh et al (f) PDE-Retinex (g) Kratz and Nishino (h) Ancuti et al (i) Ren et al (j) Zhu et al (k) PDE-IRCES (l) PA1 (m) PA2

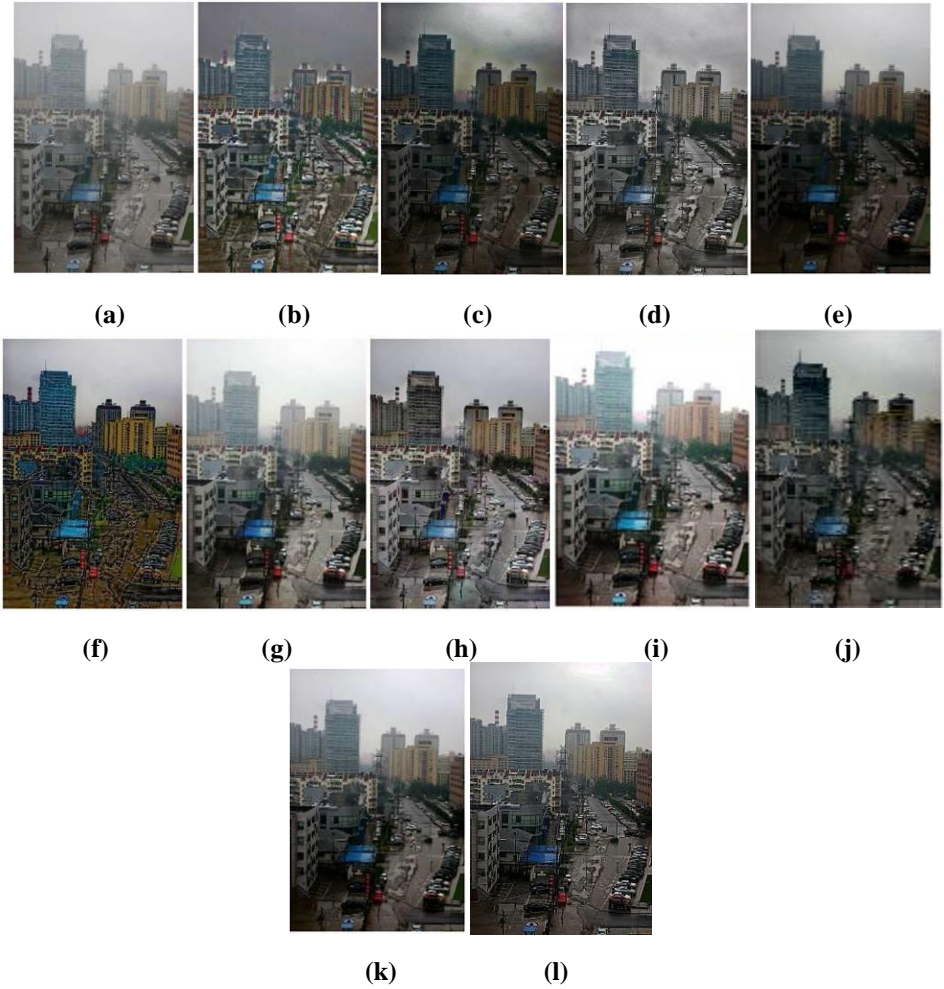

**Figure. 11** (a) Original hazy image (b) Tarel et al (c) He et al (d) PDE-Retinex (e) Zhu et al (f) Nishino, et al (g) Galdran et al (h) Wang & He (i) Dai et al (j) Ren et al (k) PDE-IRCES (l) PA

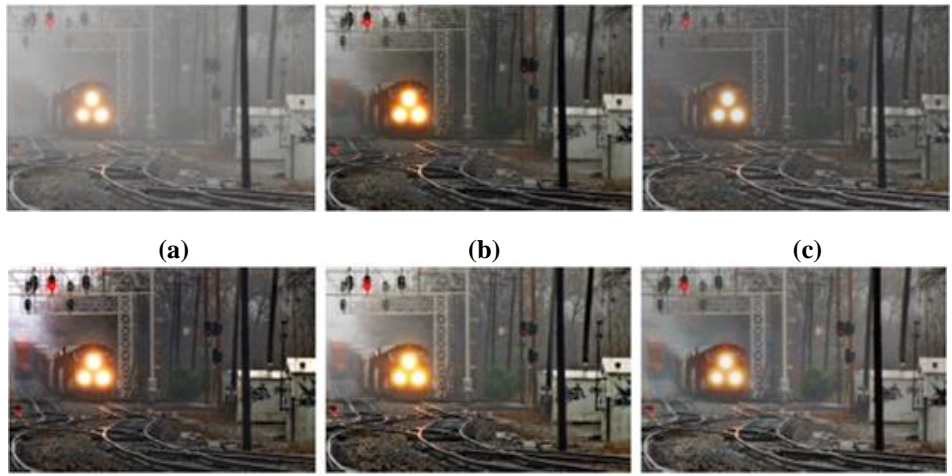

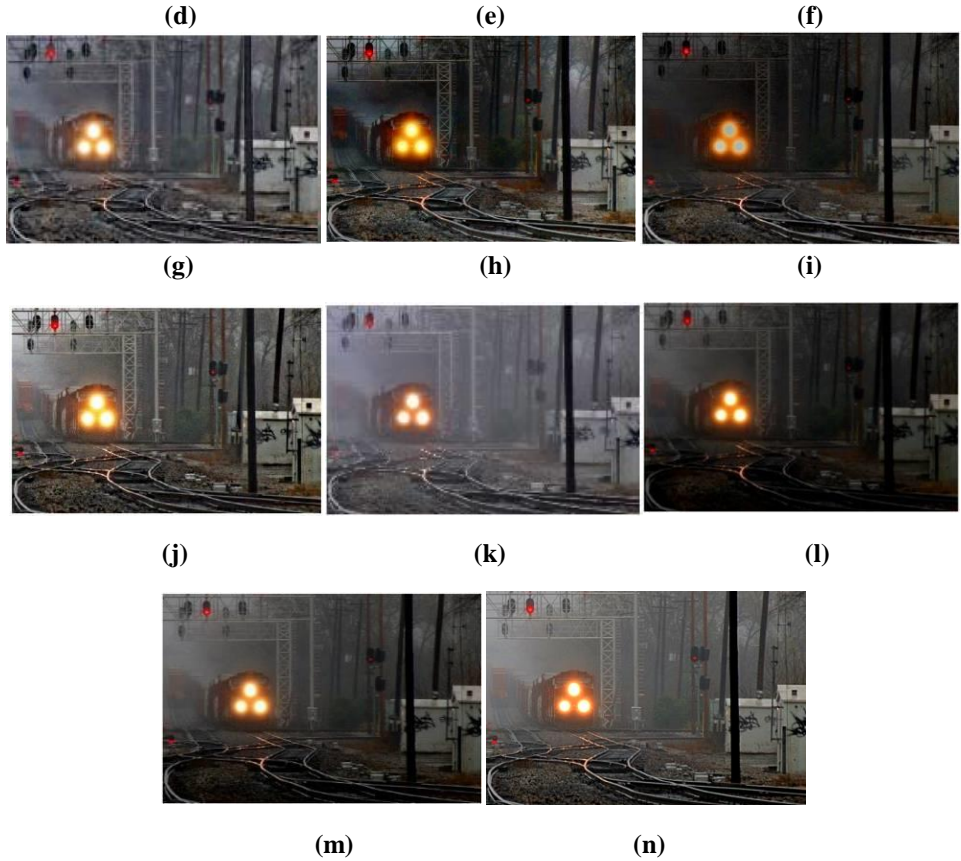

**Figure. 12** (a) Original hazy image (b) He et al (c) Tarel et al (d) Meng et al (e) Gibson et al (f) Galdran et al (g) Dai et al (h) PDE-Retinex (i) Fattal (j) Lu et al (k) Ren et al (l) Zhu et al (m) PDE-IRCES (n) PA

We also compare results of PA with those of the original DCP algorithm by He et al, which uses soft matting. The results are shown in Fig. 13 with images from the paper [1] amended with the visual results of PA. Based on the visual comparisons, PA yields comparable results with minimal runtime compared with the DCP algorithm. Thus, since the average runtime of PA is less than a second, the runtime of the original DCP algorithm is about 30 times that of the PA. This makes PA a viable option for hardware implementation for real-time operation and integration.

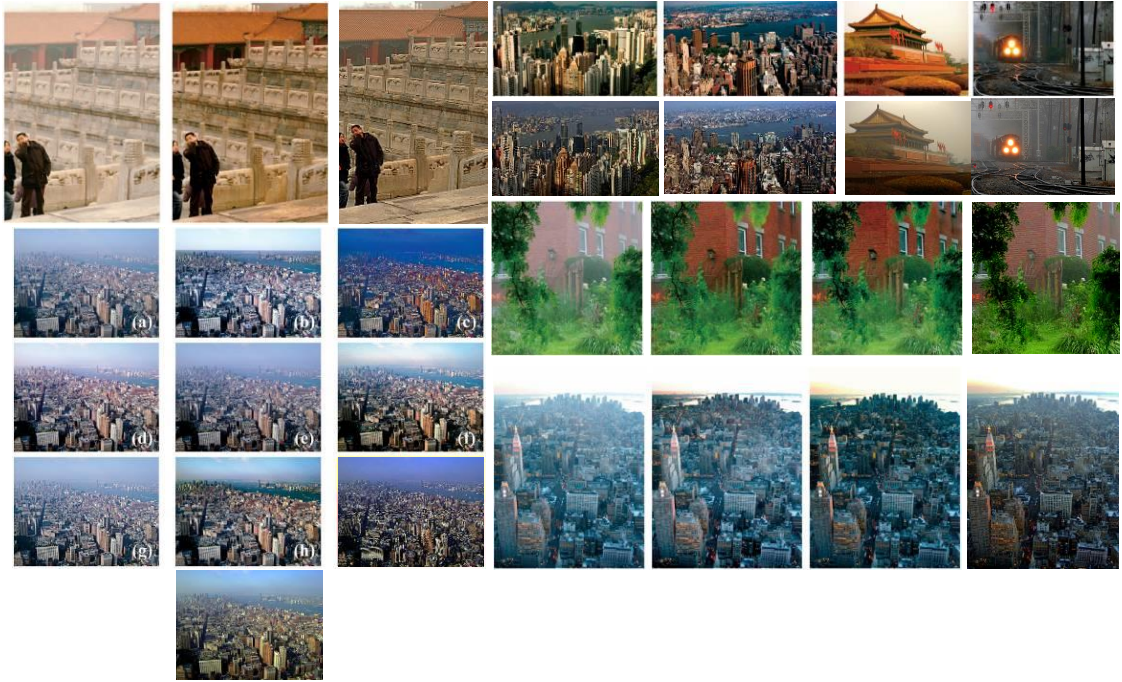

**(a)**

| hazy image (*gugong*) | He et al (Original DCP) | PA (HBFC) | | Original DCP (*hongkong*) | Original DCP (*ny1a*) | Original DCP (*Tiananmen*) | Original DCP (*Train*) |
|---|---|---|---|---|---|---|---|
| hazy image (*ny2*) | Kopf et al | Tan | | PA (*hongkong*) | PA (*ny1a*) | PA (*Tiananmen*) | PA (*Train*) |
| Fattal | Photoshop auto curve | Photoshop Unsharp Mask | | hazy image (*brickhouse*) | He et al (Original DCP) | Fattal | PA |
| Histogram Equalization | He et al (Original DCP) | PA (HPFC) | | hazy image (*ny1*) | Kopf et al | He et al (Original DCP) | PA |
| | PA (HBFC) | | | | | | |

**(b)**

**Figure 13.** (a) Visual comparison of PA with original DCP algorithm using soft matting (b) Key to figures

## 5. Conclusion

This report has presented the results of a fractional multi-scale, fusion-based hazy and underwater image enhancement algorithm. The algorithm is of greatly reduced computational complexity when compared with algorithms from the literature. The proposed approach has been compared with numerous images and algorithms from the literature and shows consistently good results.

## References

[1] Kaimin He, Jian Sun, and Xiaoou Tang, "Single Image Haze Removal Using Dark Channel Prior," *IEEE Transactions on Pattern Analysis and Machine Intelligence (PAMI)*, vol. 33, no. 12, pp. 2341-2353, 2010.

[2]  Sungmin Lee, Seokmin Yun, Ju-Hun Nam, Chee Sun Won, and Seung-Won Jung, "A review on dark channel prior based image dehazing algorithms," *EURASIP Journal on Image and Video Processing*, vol. 2016, no. 4, pp. 1-23, 2016.

[3]  Shuai Fang, Jiqing Zhan, Yang Cao, and Ruizhong Rao, "Improved single image dehazing using segmentation," in *17th IEEE International Conference on Image Processing (ICIP)*, 26-29 Sept. 2010, pp. 3589-3592.

[4]  Tong Cui, Jiandong Tian, Ende Wang, and Yandong Tang, "Single image dehazing by latent region-segmentation based transmission estimation and weighted L1-norm regularisation," *IET Image Processing* , vol. 11, no. 2, pp. 145-154, Jan. 16 2017.

[5]  Vinuchackravarthy Senthamilarasu, Anusha Baskaran, and Krishnan Kutty, "A New Approach for Removing Haze from Images," in *Proceedings of the International Conference on Image Processing, Computer Vision, and Pattern Recognition (IPCV)*, The Steering Committee of The World Congress in Computer Science, Computer Engineering and Applied Computing (WorldComp)., Jan 1, 2014, p. 1.

[6]  Codruta Orniana Ancuti, Cosmin Ancuti, and Philippe Bekaert, "Effective single image dehazing by fusion," in *17th IEEE International Conference on Image Processing (ICIP)*, 26-29 Sept. 2010, pp. 3541-3544.

[7]  Adrian Galdran, Javier Vazquez-Corral, David Pardo, and Marcelo Bertalmıo, "Fusion-based Variational Image Dehazing," *IEEE Signal Processing Letters*, vol. 24, no. 2, pp. 151-155, Feb 2017.

[8]  Peter Carr and Richard Hartley, "Improved Single Image Dehazing using Geometry," in *IEEE Digital Image Computing: Techniques and Applications, 2009. DICTA'09.* , Dec. 1, 2009 , pp. 103-110.

[9]  Dubok Park, David K. Han, and Hanseok Ko, "Single image haze removal with WLS-based edge-preserving smoothing filter," in *IEEE International Conference on Acoustics, Speech and Signal Processing (ICASSP)*, 26-31 May 2013, pp. 2469-2473.

[10]  Adrian Galdran, Javier Vazquez-Corral, David Pardo, and Marcelo Bertalmio, "A Variational Framework for Single Image Dehazing," , September 2014, pp. 1-12.

[11]  Adrian Galdran, Javier Vazquez-Corral, David Pardo, and Marcelo Bertalmio, "Enhanced Variational Image Dehazing," *SIAM Journal on Imaging Sciences*, pp. 1-26, September 2015.

[12] Xuan Liu, Fanxiang Zeng, Zhitong Huang, and Yuefeng Ji, "Single color image dehazing based on digital total variation filter with color transfer," in *20th IEEE International Conference on Image Processing (ICIP)*, 15-18 Sept. 2013, pp. 909-913.

[13] Xue-Mei Dong, Xi-Yuan Hu, Si-Long Peng, and Duo-Chao Wang, "Single color image dehazing using sparse priors," in *17th IEEE International Conference on Image Processing (ICIP)*, 26-29 Sept. 2010.

[14] Gaofeng Meng, Ying Wang, Jiangyong Duan, Shiming Xiang, and Chunhong Pan, "Efficient Image Dehazing with Boundary Constraint and Contextual Regularization," in *IEEE International Conference on Computer Vision (ICCV-2013)*, 2013, pp. 617-624.

[15] Xian-Shi Zhang, Shao-Bing Gao, Chao-Yi Li, and Yong-Jie Li, "A Retina Inspired Model for Enhancing Visibility of Hazy Images," *Frontiers in Computer Science*, vol. 9, no. 151, pp. 1-13, 22nd December 2015.

[16] Wenqi Ren et al., "Single Image Dehazing via Multi-Scale Convolutional Neural Networks," in *European Conference on Computer Vision*, Springer International Publishing, Oct. 8, 2016, pp. 154-169.

[17] Shuai Yang, Qingsong Zhu, Jianjun Wang, Di Wu, and Yaoqin Xie, "An Improved Single Image Haze Removal Algorithm Based on Dark Channel Prior and Histogram Specification," in *3rd International Conference on Multimedia Technology (ICMT-13)*, Nov. 22, 2013.

[18] Fan Guo, Zixing Cai, Bin Xie, and Jin Tang, "Automatic Image Haze Removal Based on Luminance Component," in *6th International Conference on Wireless Communications Networking and Mobile Computing (WiCOM)*, 23-25 Sept. 2010, pp. 1-4.

[19] Deepa Nair, Pattem Ashok Kumar, and Praveen Sankaran, "An Effective Surround Filter for Image Dehazing," in *ICONIAAC '14*, Amritapuri, India, October 10 - 11 2014, pp. 1-6.

[20] Bin Xie, Fan Guo, and Zixing Cai, "Improved Single Image Dehazing Using Dark Channel Prior and Multi-scale Retinex," in *International Conference on Intelligent System Design and Engineering Application (ISDEA)*, 13-14 Oct. 2010, pp. 848-851, Vol. 1.

[21] U. A. Nnolim. (2017) Sky detection and log illumination refinement for PDE-based hazy image contrast enhancement. [Online]. http://arxiv.org/pdf/1712.09775.pdf

[22] U. A. Nnolim, "Partial differential equation-based hazy image contrast enhancement," *Computers and Electrical Engineering*, vol. (in print), pp. 1-11, February 21 2018.

[23] U. A. Nnolim, "Image de-hazing via gradient optimized adaptive forward-reverse flow-based partial differential equation," *Journal of Circuits Systems and Computers*, no. (accepted), pp. 1-35, July 5 2018.

[24] Adrian Galdran, "Artificial Multiple Exposure Image Dehazing," *Signal Processing*, vol. 149, pp. 135-147, August 2018.

[25] He B, Wu Q Zhu M, "Single Image Dehazing Based on Dark Channel Prior and Energy Minimization," *IEEE Signal Processing Letters*, vol. 25, no. 2, pp. 174-178, Feb 2018.

[26] Zhu M, Xia Z, Zhao M Shi Z, "Fast single-image dehazing method based on luminance dark prior," *International Journal of Pattern Recognition and Artificial Intelligence*, vol. 31, no. 2, p. 1754003, Feb. 2017.

[27] X. Yuan, M. Ju, Z. Gu, and S. Wang, "An Effective and Robust Single Image Dehazing Method Using the Dark Channel Prior," *Information* , vol. 8, no. 57, 2017.

[28] Tang G, Zhang X, Jiang J, Tian Q Zhu Y, "Haze removal method for natural restoration of images with sky," *Neurocomputing*, vol. 275, pp. 499-510, Jan 31 2018.

[29] Ju M, Zhang D. Wang X, "Automatic hazy image enhancement via haze distribution estimation," *Advances in Mechanical Engineering.*, vol. 10 , no. 4, p. 1687814018769485, Apr 2018.

[30] Li X Du Y, "Recursive Deep Residual Learning for Single Image Dehazing," in *Proceedings of the IEEE Conference on Computer Vision and Pattern Recognition Workshops 2018*, 2018, pp. 730-737.

[31] Lu N, Yao L, Zhang X. Jiang H, "Single image dehazing for visible remote sensing based on tagged haze thickness maps," *Remote Sensing Letters*, vol. 9, no. 7, pp. 627-635, July 3 2018.

[32] Ding C, Zhang DY, Guo YJ Ju MY, "Gamma-Correction-Based Visibility Restoration for Single Hazy Images," *IEEE Signal Processing Letters*, May 22 2018.

[33] Sim H, Choi JS, Seo S, Kim S, Kim M Ki S, "Fully End-to-End learning based Conditional Boundary Equilibrium GAN with Receptive Field Sizes Enlarged for Single Ultra-High Resolution Image Dehazing," in *Proceedings of the IEEE Conference on Computer Vision and Pattern Recognition Workshops 2018*, 2018, pp. 817-824.

[34] Chongyi Li, Jichang Guo, Fatih Porikli, Huazhu Fu, and Yanwei Pang, "A Cascaded Convolutional Neural Network for Single Image Dehazing," *IEEE Access*, vol. 6, pp. 24877-24887, March 23, 2018.

[35] Pan J, Li Z, Tang J Li R, "Single Image Dehazing via Conditional Generative Adversarial Network. ," *Methods*, vol. 3, no. 24, 2018.

[36] Zeng H, Shang Y, Shao Z, Ding H. Luan Z, "Fast Video Dehazing Using Per-Pixel Minimum Adjustment," *Mathematical Problems in Engineering*, 2018.

[37] Santra S, Chanda B Mondal R, "Image Dehazing by Joint Estimation of Transmittance and Airlight using Bi-Directional Consistency Loss Minimized FCN," in *Proceedings of the IEEE Conference on Computer Vision and Pattern Recognition Workshops 2018*, pp. 920-928.

[38] Xie F, Li W, Shi Z, Zhang H Qin M, "Dehazing for Multispectral Remote Sensing Images Based on a Convolutional Neural Network With the Residual Architecture," *IEEE Journal of Selected Topics in Applied Earth Observations and Remote Sensing*, vol. 11, no. 5, pp. 1645-1655, May 2018.

[39] Mondal R, Chanda B Santra S, "Learning a Patch Quality Comparator for Single Image Dehazing," *IEEE Transactions on Image Processing*, vol. 27, no. 9, May 28 2018.

[40] Li J, Wang X, Chen X. Song Y, "Single Image Dehazing Using Ranking Convolutional Neural Network," *IEEE Transactions on Multimedia*, vol. 20, no. 6, pp. 1548-1560, June 2018.

[41] Sindagi V, Patel VM Zhang H, "Multi-scale Single Image Dehazing using Perceptual Pyramid Deep Network," in *Proceedings of the IEEE Conference on Computer Vision and Pattern Recognition Workshops 2018*, pp. 902-911.

[42] Li G, Fan H Li J, "Image Dehazing using Residual-based Deep CNN," *IEEE Access*, May 7 2018.

[43] Jingyu Lu et al., "Multi-scale adversarial network for underwater image restoration," *Optics and Laser Technology*, no. (article in press), 2018.

[44] A. Galdran, D. Pardo, A. Picón, and A. Alvarez-Gila, "Automatic red-channel underwater image restoration," *Journal of Visual Communication and Image Representation*, vol. 26, pp. 132-145, Jan 31 2015.

[45] Chongyi Li et al., "Single underwater image enhancement based on color cast removal and visibility restoration," *Journal of Electronic Imaging*, vol. 25, no. 3, pp. 1-15, June 2016.

[46] C. Li and J. Guo, "Underwater image enhancement by dehazing and color correction," *SPIE Journal of Electronic Imaging*, vol. 24, no. 3, p. 033023, 22 June 2015.

[47] X. Zhao, T. Jin, and S. Qu, "Deriving inherent optical properties from background color and underwater image enhancement," *Ocean Engineering*, vol. 94, pp. 163-172, 2015.

[48] J. Chiang and Y. Chen, "Underwater image enhancement by wavelength compensation and dehazing," *IEEE Transactions on Image Processing*, vol. 21, no. 4, pp. 1756-1769, 2012.

[49] Haocheng Wen, Yonghong Tian, Tiejun Huang, and Wen Gao, "Single underwater image enhancement with a new optical model," in *IEEE International Symposium on Conference on Circuits and Systems (ISCAS)*, May 2013, pp. 753-756.

[50] S. Serikawa and H. Lu, "Underwater image dehazing using joint trilateral filter," *Computers in Elecrical Engineering*, vol. 40, no. 1, pp. 41-50, 2014.

[51] N. Carlevaris-Bianco, A. Mohan, and R. M. Eustice, "Initial results in underwater single image dehazing," in *Proceedings of IEEE International Conference on Oceans*, 2010, pp. 1-8.

[52] John Y. Chiang, Ying-Ching Chen, and Yung-Fu Chen, "Underwater Image Enhancement: Using Wavelength Compensation and Image Dehazing (WCID)," in *ACIVS 2011, LNCS 6915*, 2011, pp. pp. 372–383.

[53] K. Iqbal, R. Abdul Salam, A. Osman, and A Zawawi Talib, "Underwater image enhancement using an integrated color model," *IAENG International Journal of Computer Science*, vol. 34, no. 2, pp. 529-534, January 2007.

[54] A. S. A. Ghani and N. A. M. Isa, "Underwater image quality enhancement through integrated color model with Rayleigh distribution," *Applied Soft Computing*, vol. 27, pp. 219-230, 2015.

[55] Xueyang Fu et al., "A retinex-based enhancing approach for single underwater image," in *Proceedings of International Conference on Image Processing (ICIP)*, 27 October 2014, pp. 4572-4576.

[56] H. Gouinaud, Y. Gavet, J. Debayle, and J.-C. Pinoli, "Color Correction in the Framework of Color Logarithmic Image Processing," in *IEEE 7th International Symposium on Image and Signal Processing and Analysis (ISPA 2011)*, Dubrovnik, Croatia, Sep 2011, pp. 129-133.

[57] S. Bazeille, I. Quidu, L. Jaulin, and J. P. Malkasse, "Automatic underwater image pre-processing," in *Proceedings of the Characterisation du Milieu Marin (CMM '06)*, 2006.

[58] M. Chambah, D. Semani, Arnaud Renouf, P. Coutellemont, and A. Rizzi, "Underwater Color Constancy : Enhancement of Automatic Live Fish Recognition," in *16th Annual symposium on Electronic Imaging*, Inconnue, United States, 2004, pp. 157-168.

[59] L. A. Torres-Mendez and G Dudek, "Color correction of underwater images for aquatic robot inspection," in *Proceedings of the 5th International Workshop on Energy Minimization Methods in Computer Vision and Pattern Recognition (EMMCVPR '05)*, Augustine, Fla, USA, November 2005, pp. vol. 3757, pp. 60–73.

[60] J. Ahlen, D. Sundgren, and E. Bengtsson, "Application of underwater hyperspectral data for color correction purposes," *Pattern Recognition and Image Analysis*, vol. 17, no. 1, pp. 170–173, 2007.

[61] Julia Ahlen, *Colour correction of underwater images using spectral data*.: Uppsala University, 2005.

[62] F Petit, A-S Capelle-Laizé, and P Carré, "Underwater image enhancement by attenuation inversion with quaternions," in *Proceedings of the IEEE International Conference on Acoustics, Speech and Signal Processing (ICASSP '09)*, Taiwan, 2009, pp. pp. 1177–1180.

[63] G. Bianco, M. Muzzupappa, F. Bruno, R. Garcia, and L. Neumann, "A New Colour Correction Method For Underwater Imaging," *The International Archives of the Photogrammetry, Remote Sensing and Spatial Information Sciences Underwater 3D Recording and Modeling*, vol. XL-5/W5, no. 5, Piano di Sorrento, Italy, pp. 25-32, 16–17 April 2015.

[64] C.J. Prabhakar and P.U. Kumar Praveen, "An Image Based Technique for Enhancement of Underwater Images," *International Journal of Machine Intelligence*, vol. 3, no. 4, pp. 217-224, 2011.

[65] Huimin Lu, Yujie Li, Lifeng Zhang, and Seiichi Serikawa, "Contrast enhancement for images in turbid water," *Journal of Optical Society of America*, vol. 32, no. 5, pp. 886-893, May 1 2015.

[66] U. A. Nnolim, "Smoothing and enhancement algorithms for underwater images based on partial differential equations," *SPIE Journal of Electronic Imaging*, vol. 26, no. 2, pp. 1-21, March 22 2017.

[67] Uche A. Nnolim, "Improved partial differential equation (PDE)-based enhancement for underwater images using local-global contrast operators and fuzzy homomorphic processes," *IET Image Processing*, vol. 11, no. 11, pp. 1059-1067, November 2017.

[68] R. Garcia, T. Nicosevici, and X. Cufi, "On the way to solve lighting problems in underwater imaging," in *Proceedings of the IEEE Oceans Conference Record*, 2002, pp. vol. 2, pp. 1018–1024.

[69] Y. Rzhanov, L. M. Linnett, and R. Forbes, "Underwater video mosaicing for seabed mapping," in *Proceedings of IEEE International Conference on Image Processing*, 2000, pp. vol. 1, pp. 224–227.

[70] H. Singh, J. Howland, D. Yoerger, and L. Whitcomb, "Quantitative photomosaicing of underwater imaging," in *Proceedings of the IEEE Oceans Conference*, 1998, pp. vol. 1, pp. 263–266.

[71] U. A. Nnolim, "Adaptive multi-scale entropy fusion de-hazing based on fractional order," *Preprints*, 2018.

[72] C. Ancuti, CO. Ancuti, T. Haber, and P. Bekaert, "Enhancing underwater images and videos by fusion," in *IEEE Conference on Computer Vision and Pattern Recognition*, Jun 16 2012, pp. 81-88.

[73] Sabine Susstrunk and David Hasler, "Measuring Colourfulness in Natural Images," in *IS&T/SPIE Electronic Imaging 2003: Human Vision and Electronic Imaging VIII*, 2003, pp. vol. 5007, pp. 87-95.

[74] Kresimir Matkovic, Laszlo Neumann, Attila Neumann, Thomas Psik, and Werner Purgathofer, "Global Contrast Factor-a New Approach to Image Contrast," in *Proceedings of the 1st Eurographics Conference on Computational Aesthetics in Graphics, Visualization and Imaging*, Girona, Spain, May 18-20, 2005, pp. 159-167.

[75] Miao Yang and Arcot Sowmya, "An Underwater Color Image Quality Evaluation Metric," *IEEE Transactions on Image Processing*, vol. 24, no. 12, pp. 6062-6071, December 2015.

[76] Simon Emberton, Lars Chittka, and Andrea Cavallaro, "Underwater image and video dehazing with pure haze region segmentation," *Computer Vision and Image Understanding* , vol. 168, pp. 145–156, 2018.

[77] K. Panetta, C. Gao, and S. Agaian, "Human-visual-system-inspired underwater image quality measures," *IEEE Journal of Ocean Engineering*, vol. 41, no. 3, pp. 541-551, 2016.

[78] Qingsong Zhu, Jiaming Mai, and Ling Shao, "A Fast Single Image Haze Removal Algorithm Using Color Attenuation Prior," *IEEE Transactions on Image Processing*, vol. 24, no. 11, pp. 3522-3533, November 2015.

[79] J.P. Tarel and N. Hautiere, "Fast visibility restoration from a single color or gray level image," in *IEEE 12th International Conference on Computer Vision*, Sep 2009, pp. 2201-2208.

[80] Wei Wang and Chuanjiang He, "Depth and Reflection Total Variation for Single Image Dehazing," College of Mathematics and Statistics, Chongqing University, Chongqing, China, Technical report 1601.05994, 13 October October 2016.

[81] Sheng-kui Dai and Jean-Philippe Tarel, "Adaptive Sky Detection and Preservation in Dehazing Algorithm," in *IEEE International Symposium on Intelligent Signal Processing and Communication Systems (ISPACS)*, November 2015, pp. 634-639.

[82] K. Nishino, L. Kratz, and S. Lombardi, "Bayesian Defogging," *International Journal of Computer Vision*, vol. 98, no. 3, pp. 263–278, July 2012.